\documentclass{article}

\usepackage{arxiv}

\usepackage[utf8]{inputenc} 
\usepackage[T1]{fontenc}    
\usepackage{hyperref}       
\usepackage{url}            
\usepackage{booktabs}       
\usepackage{amsfonts}       
\usepackage{nicefrac}       
\usepackage{microtype}      
\usepackage{lipsum}		
\usepackage{graphicx}
\usepackage{natbib}
\usepackage{doi}
\usepackage{multirow}
\usepackage{longtable}
\usepackage{geometry}
\usepackage{booktabs}
\usepackage{subcaption}

\title{An Explainable Machine Learning Approach to Traffic Accident Fatality Prediction}

\author{ \href{https://orcid.org/0009-0006-7852-915X}{\includegraphics[scale=0.06]{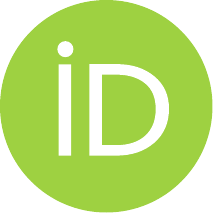}\hspace{1mm}Md. Asif Khan Rifat} \\
	Institute of Information Technology\\
	University of Dhaka\\
	Dhaka-1000, Bangladesh \\
	\texttt{msse1728@iit.du.ac.bd} \\
	\And
	\href{https://orcid.org/0000-0001-5983-6775}{\includegraphics[scale=0.06]{orcid.pdf}\hspace{1mm}Ahmedul Kabir} \thanks{Corresponding Author.} \\
	Institute of Information Technology\\
	University of Dhaka\\
	Dhaka-1000, Bangladesh \\
	\texttt{kabir@iit.du.ac.bd} \\ 
        \And
	\href{https://orcid.org/0000-0002-1719-8194}{\includegraphics[scale=0.06]{orcid.pdf}\hspace{1mm}Armana Sabiha Huq} \\
	Accident Research Institute\\
	Bangladesh University of Engineering and Technology\\
	Dhaka-1000, Bangladesh \\
	\texttt{ashuq@ari.buet.ac.bd} \\        
}

\renewcommand{\shorttitle}{\textit{Rifat et al.}}

\hypersetup{
pdftitle={A template for the arxiv style},
pdfsubject={q-bio.NC, q-bio.QM},
pdfauthor={David S.~Hippocampus, Elias D.~Striatum},
pdfkeywords={First keyword, Second keyword, More},
}

\begin{document}
\maketitle

\begin{abstract}
Road traffic accidents (RTA) pose a significant public health threat worldwide, leading to considerable loss of life and economic burdens. This is particularly acute in developing countries like Bangladesh. Building reliable models to forecast crash outcomes is crucial for implementing effective preventive measures. To aid in developing targeted safety interventions, this study presents a machine learning-based approach for classifying fatal and non-fatal road accident outcomes using data from the Dhaka metropolitan traffic crash database from 2017 to 2022. Our framework utilizes a range of machine learning classification algorithms, comprising Logistic Regression, Support Vector Machines, Naive Bayes, Random Forest, Decision Tree, Gradient Boosting, LightGBM, and Artificial Neural Network.  We prioritize model interpretability by employing the SHAP (SHapley Additive exPlanations) method, which elucidates the key factors influencing accident fatality. Our results demonstrate that LightGBM outperforms other models, achieving a ROC-AUC score of 0.72. The global, local, and feature dependency analyses are conducted to acquire deeper insights into the behavior of the model. SHAP analysis reveals that casualty class, time of accident, location, vehicle type, and road type play pivotal roles in determining fatality risk. These findings offer valuable insights for policymakers and road safety practitioners in developing countries, enabling the implementation of evidence-based strategies to reduce traffic crash fatalities.
\end{abstract}


\section{Introduction}
\label{main}

Injuries caused by road traffic crashes are one of the leading reasons of death every year all over the globe. As per the Centers for Disease Control and Prevention (\citet{CDC}), road accident injuries rank as the eighth most prevalent cause of death worldwide across all age groups and are projected to be the primary cause of death among individuals aged 5 to 29 years according to. Reportedly, someone is being killed every 24 seconds due to a road accident \citep{bloomberg_global}. Due to road traffic crashes, not only people's lives are being cut short each year, but almost 20-50 million more people suffer non-fatal injuries, among them many incur disabilities as well. The World Health Organization (\citet{WHO}) estimates that 92\% of the fatalities on the road occur in low and middle-income countries. Road accidents in a developing country such as Bangladesh have become a significant concern, leading to a high number of casualties and economic losses each year. It has been found that in Bangladesh, one of the reasons behind this substantial economic burden is that 67\% of the fatalities caused by road crashes occur within the economically productive age group of 15-64 years \citep{bangladeshs}. According to the Bangladesh Road Safety Foundation’s (\citet{RSF}) annual report, at least 6,284 people died, and 7,468 others were injured in road accidents in 2021, compared to 5,431 people dead and 7,379 injured in road collisions in 2020. Given the huge number of vehicles in the complex transportation system, we need not only an understanding of the causes of accidents but also a reliable and precise traffic accident outcome prediction model to mitigate potential traffic accidents.

The alarming global situation regarding the substantial consequences of road traffic crashes serves as a constant reminder for us to implement effective measures for preventing such incidents. This phenomenon has led us to prioritize the exploration of the causes behind road traffic crashes. Many research works in the past have employed different methods and techniques to understand different aspects of road crashes and their consequences. Road crashes result in different levels of injury severity, and various contributing factors exert varying degrees of influence on the severity of these injuries. Researchers have increasingly concentrated on building various models for forecasting the fatality and severity of injuries in road traffic incidents as the frequency of road crashes has increased over the last several decades. Traditionally, these models relied on mathematical and statistical analysis, but more recently, Artificial Intelligence (AI), specifically Machine Learning (ML) algorithms, are being used to anticipate various crash-causing events and classify the crash injury severity. Integration of explainable AI (XAI) methods like SHAP and LIME along with predictive models is an effective way to explore the crash-influencing factors \citep{dong_2022_predicting}. Road accidents are multifaceted events influenced by a myriad of factors, including characteristics of the road, vehicles involved, human behaviors, and speed regulations. Establishing a direct or linear relationship between fatal and non-fatal cases is challenging due to the complexity of these incidents, as multiple factors often converge \citep{Ahmed_2023}. This complexity presents a significant barrier to the development of effective road safety models. 

In this study, our objective is to predict individual road crash fatalities for all vehicle occupants, including drivers, passengers, and pedestrians, and to build a framework that identifies the most informative variables for distinguishing fatal from non-fatal road accidents, with a specific focus on the context of Bangladesh. To achieve this goal, we have employed various machine learning models, including Logistic Regression, Support Vector Machines (SVM), Naive Bayes, Random Forest, Decision Tree, Gradient Boosting and LightGBM. Additionally, we have utilized a Neural Network model to predict crash fatalities. Our analysis is based on data sourced from the Dhaka Metropolitan Police (DMP) traffic accident database spanning from 2017 to 2022. Toward the conclusion of our study, we employed the SHAP (Shapley Additive exPlanations) method for model interpretability, which enables a deeper understanding of the specific crash-influencing factors pertinent to Dhaka city. The prediction of crash fatalities and the interpretability of our predictive algorithms hold significant promise in informing strategies for enhancing road traffic safety within the context of a developing country.  

\section{Literature Review}
Road traffic accidents pose significant risks to public safety, with fatalities being a primary concern. Understanding the factors influencing accident severity and fatality is crucial for implementing effective preventive measures. This literature review consolidates previous research works in analyzing and predicting accident outcomes, incorporating both statistical methodologies and machine learning approaches.

\subsection{Statistical and ML Modeling of Accident Outcomes}
Earlier studies, such as \citet{Abdel_Aty_2004} compared two distinct paradigms of Artificial Neural Networks (ANN) and the fuzzy Adaptive Resonance Theory (ART) for predicting the severity level of a crash. The study revealed that ANN was more effective in predicting the severity level of a crash as compared to the calibrated ordered probit model. Subsequent research showed a shift towards machine learning techniques for classification. Factors affecting the severity of pedestrian accidents were discussed by \citet{Haleem_2015}. The authors of this study used Random Forest to determine the most important parameters and a mixed logit model to predict the severity. \citet{Kabeer_2016} proposed the use of an ensemble technique to analyze road accidents in Leeds. It is shown that ensemble techniques can increase the prediction accuracy to 78.03\% while the accuracy
of Naïve Bayes and DTs was 58.76\% and 51.22\%, respectively. 
A combination of clustering and classification is used by \citet{Iranitalab_2017} to improve classification accuracy, while \citet{mashfiq2021data} employed data mining for identifying accident-prone areas, offering valuable insights for policymakers. Recent studies have increasingly utilized machine learning for crash severity prediction. \citet{Ijaz_2021} presented a comparative study of Decision Tree, and Random Forest for fatality prediction involving crashes of three-wheeled motorized rickshaws. Besides classical ML models, \citet{rahim2021deep} applied Convolutional Neural Network (CNN) with customized loss function for the prediction of crash severity, demonstrating the efficacy of deep learning. However, these works lack explainability due to the opaque nature of black box models, which deprives the understanding of contributing features and their dependencies on the model's output.

\subsection{Interpretable Models and Factors Analysis}
Several studies focused on developing interpretable models to understand factors contributing to injury severity. \citet{dong_2022_predicting} employed boosting-based ensemble learning models, highlighting significant variables such as month and driver age. \citet{jpssashiprabhamadushani_2023_evaluating} investigated crash severity factors in Sri Lanka using explainable machine learning methods, emphasizing their superiority over traditional regression analysis. \citet{eboli_2020_factors} utilized logistic regression to dissect crash-contributing factors, providing practical implications to formulate road safety measures. \citet{li_2017_analysis} analyzed fatal accidents using data mining techniques, emphasizing the role of human factors like intoxication. \citet{rocha_2022_identifying} introduced a framework for identifying informative variables for distinguishing fatal from non-fatal accidents, offering valuable insights for accident prevention strategies. Dealing with imbalanced data remains a challenge in crash severity classification. \citet{jeong_2018_classification} addressed this challenge by employing various classification models and data treatment techniques. \citet{wen_2021_applications} proposed solutions for handling imbalanced crash severity data, advocating for interpretable machine learning models and advanced techniques like graph convolutional networks. Study of \citet{Rekha2023} focuses on using neural networks to predict factors influencing fatality rates in road accidents. It addresses the challenge of class imbalance and emphasizes the relationship between attributes and fatality rates in accidents.

The reviewed literature highlights the evolution from statistical models to machine learning approaches in analyzing and predicting accident severity. Interpretable models provide insights into factors influencing crash outcomes, aiding in the formulation of effective safety policies. Addressing challenges like imbalanced data and model ambiguity remains crucial for enhancing the accuracy and applicability of predictive models in traffic safety management. 

\section{Methods}
\begin{figure}[!ht]
  \centering
  \includegraphics[width=0.65\textwidth]{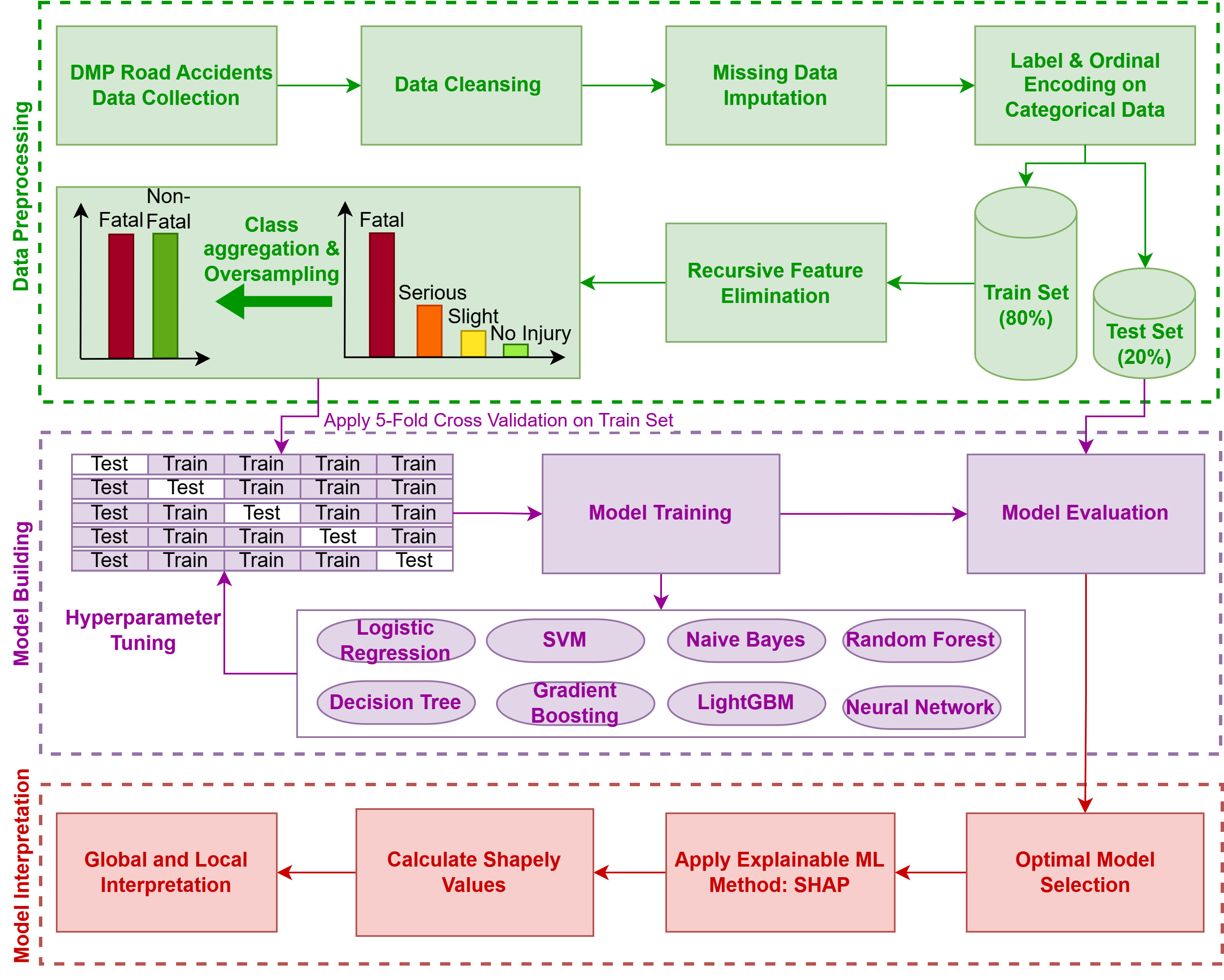}
  \caption{Framework of road accident fatality prediction with model interpretability using SHAP XAI.}\label{fig:fig1}
\end{figure}

The study mainly concerns the binary classification of road accident fatality. It involves data collection, preprocessing, feature engineering, model development, evaluation, interpretation, and
analysis. The proposed methodology diagram of the study is shown in Figure~\ref{fig:fig1}.

\subsection{Dataset Overview and Preprocessing}
\label{data}
The dataset was collected from Dhaka Metropolitan Police (DMP) which contains traffic accident records from 2017-2022 in Dhaka, Bangladesh. It consists of 50 attributes and over 2000 samples detailing accident locations, environment, time, road characteristics, vehicle data, human behavior, etc. The raw data undergoes preprocessing, which involves removing variables with excessive missing values, filling in missing data, and encoding categorical variables. Columns irrelevant to accident fatality, such as license number and vehicle registration were dropped, and instances with null values were eliminated. New columns based on domain knowledge were generated, such as casualty class and indicators for weekends and times of day. Exploratory data analysis revealed class imbalance in the target class, accident severity, originally encompassed four types: fatal, serious, slight, and no injury. Particularly slight and no-injury instances constitute a marginal frequency of 0.07\% of the total. This skew is atypical compared to other developed countries where most accidents tend to be fatal. To address this class imbalance, serious, slight, and no-injury classes were combined into "non-fatal," and "fatal" includes records with at least one death, resulting in 535 non-fatal and 1165 fatal instances. While this reduces the class imbalance ratio, the issue persists. To mitigate this, Synthetic Minority Over-sampling Technique (SMOTE) was applied to the training set, initially with 428 non-fatal and 931 fatal cases. SMOTE oversampling generated synthetic samples to increase the minority class and balance the dataset \citep{Smote}, resulting in a training set of 1862 cases with an equal distribution of 931 cases per category. The dataset includes numerical, categorical, and ordinal text attributes. Features with categorical variables are encoded into numerical ones using \textit{LabelEncoder}. Conversely, casualty class and accident fatality features are encoded using \textit{OrdinalEncoder}, as the order of these features is significant. To ensure uniformity and comparability among features, Min-Max Scaling is applied which rescales the features to a specified range, typically between 0 and 1. After preprocessing, the dataset comprises 1700 accident records with 33 feature variables and one target variable, spanning seven categories: traffic attributes, temporal attributes, road attributes, vehicle characteristics, driver-related factors, environmental factors, and accident outcomes listed in Table~\ref{tab:tab1}

\begin{table}[ht]
\caption{Dhaka metropolitan traffic accident dataset selected features overview.}\label{tab:tab1}
\begin{tabular*}{\textwidth}{@{\extracolsep{\fill}}p{2.25cm}p{2.25cm}p{9.5cm}@{}}
\toprule
\textbf{Feature Category} & \textbf{Feature Name} & \textbf{Description} \\
\midrule
Traffic Attributes & Sub-district & Accident occurring police stations or sub-districts in Dhaka city \\
 & Traffic control & Police Controlled, Uncontrolled, Others, Road Divider, Police and Traffic Light, Pedestrian Crossing, Traffic Lights, Traffic Sign \\
 & Collision type & Hit Pedestrian, Rear End, Side Impact, Direct Collision, Hit Parked Vehicle, Hit Object beside Road, Hit Object on Road, Overturned, Right Angle, Others \\
\midrule
Temporal Attributes & Day of week & 0-6 [Sunday-Saturday] \\
 & Weekend & 1, 0 [Yes, No] \\
 & Time & 0-23 [Hours] \\
 & Month & 0-11 [January-December] \\
 & Year & 2017-2022 [Year] \\
 & Date in month & 0-31 [Day] \\
\midrule
Road Attributes & Road class & City, National, Regional, Feeder, Rural \\
 & Divider & Yes, No \\
 & Junction type & No Junction, Cross Roads, T-Junction, Roundabout, Staggered Junction, Unknown, Railway Crossing, Others \\
 & Movement & One Way, Two Way \\
 & Surface quality &  Good, Rough, Under Maintenance \\ 
\midrule
Vehicle & Vehicle type & Bus, Heavy Truck, Others, Minibus, Car, Motorcycle, Pickup, Microbus, Mini Truck, CNG, Jeep, Tempo, Bicycle, Rickshaw/Van, Articulated Truck, Tanker, Handcart \\
 & Vehicle damage & None, Front, Right, Multiple, Unknown, Behind, Left, Roof \\
 & Vehicle maneuver & Straight, Overtaking, Others, Left-turn, Transverse Crossing, Right-turn, Backward, U-turn, Brake/Slowing Down, Sudden Acceleration, Parked \\
 & Fitness certificate & Yes, No, Unknown \\
\midrule
Driver & Driver age & Age of the driver [14-80]\\
 & Seat belt & Driver wears seat belt or not [Yes, No]\\
\midrule
Environment & Light & Daylight, Illuminated at Night, Twilight, Unlit at Night \\
  & Weather & Good, Rainy, Fog, Storm \\
\midrule
Accident Outcomes & Casualty class & Driver, Pedestrian, Passenger, Driver and Passenger, Driver and Pedestrian, Passenger and Pedestrian, All, None \\
 & Total Casualties & Total number of casualties in an accident [0-17] \\
 & Accident Fatality & Fatal, Non Fatal \\
\bottomrule
\end{tabular*}
\end{table}

\subsection{Feature Selection}
\label{fet_sel}
\begin{figure}[!ht]
  \centering
  \includegraphics[width=0.45\textwidth]{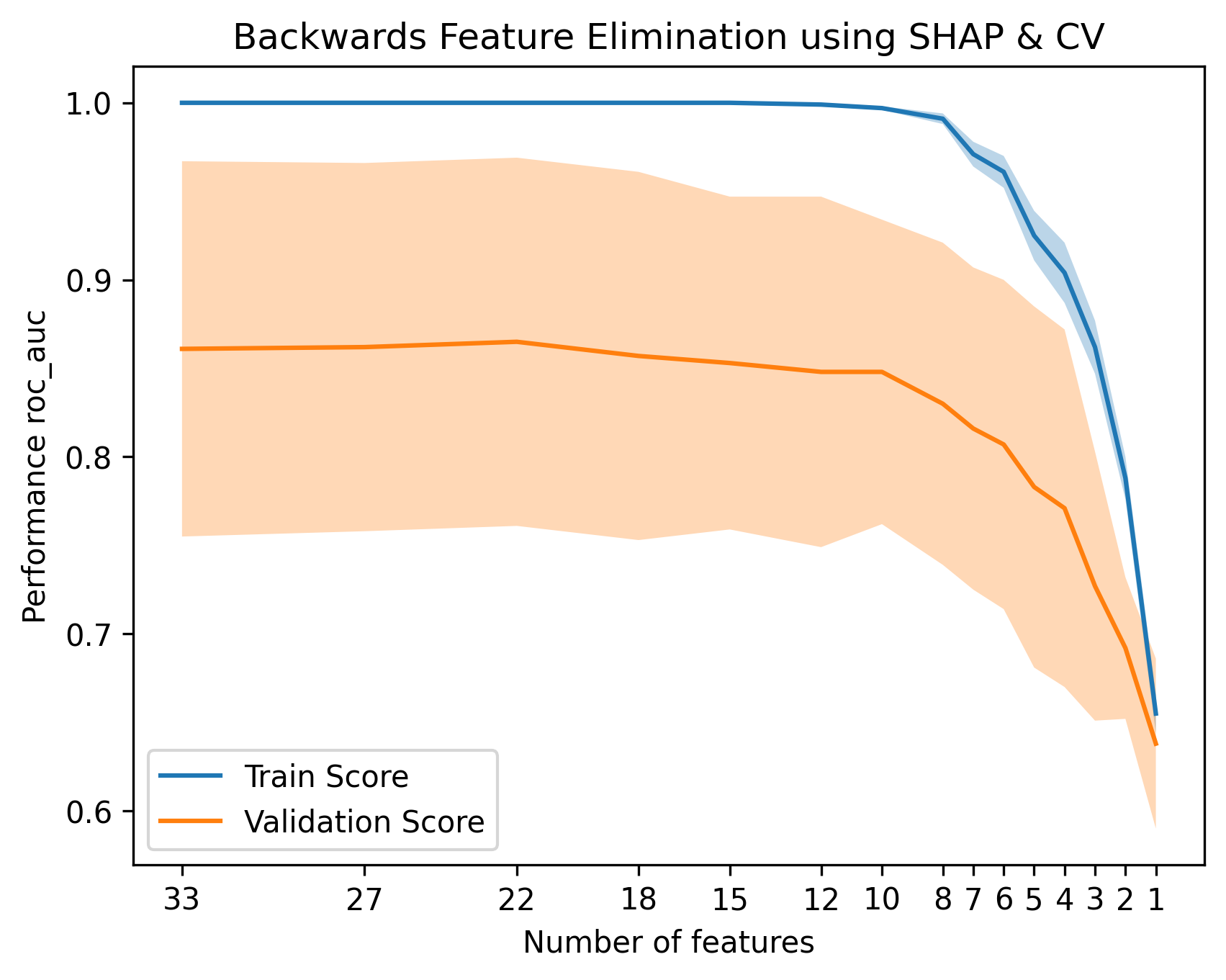}
  \caption{Recursive feature elimination with SHAP (ShapRFECV) performance visualization.}\label{fig:fig2}
\end{figure}

Given the presence of 33 dependent features, employing feature selection becomes crucial to enhance model performance. In this study, Recursive Feature Elimination with SHAP (ShapRFECV) was utilized as the feature selection technique. Backwards feature elimination is a technique used to identify the most important features in a machine learning model. It starts with a full set of features and iteratively removes the least important feature until a stopping criterion is met. The SHAP (SHapley Additive exPlanations) values are used to determine the importance of each feature. SHAP values explain how much each feature contributes to a particular prediction. The features with the lowest SHAP values are removed first. The 5-fold cross-validation (CV) was used to evaluate the performance of the model after each feature removal. The model was trained on a subset of the train set and then evaluated on a different subset of the validation set. This process was repeated multiple times using different folds of the train data. The mean ROC-AUC score on the validation set was used to assess the impact of removing a feature. By systematically training the model on various combinations of features, it becomes apparent that the model achieved optimal validation ROC-AUC score when utilizing the top 23 features as shown in Figure~\ref{fig:fig2}. Therefore, after careful analysis and validation, the decision was made to select these top 23 features for further analysis and model development. 

\subsection{Model Training}
The preprocessed dataset was divided into a training set (80\%) and a test set (20\%) with stratified random sampling. The training set was used for model development, while the held-out test set served as an unbiased benchmark for performance evaluation. Subsequently, a variety of traditional machine learning classification algorithms were employed as foundational predictive models for traffic crash fatality prediction. These include Random Forest Classifier, Logistic Regression, Naive Bayes, K Nearest Neighbour Classifier, Light Gradient Boosting, and Extreme Gradient Boosting Classifier. Additionally, a Multi-Layer Perceptron (MLP) algorithm is utilized as a neural network model, incorporating dense and dropout layers to capture complex nonlinear relationships within the data. The models were trained using the \textit{scikit-learn} and \textit{TensorFlow} libraries, with cross-validation and regularization techniques employed to mitigate overfitting and address class imbalance issues. Hyperparameter tuning was conducted using the \textit{gridsearchcv} library, optimizing the parameters of all machine learning models to maximize accuracy and improve predictive performance.

\subsection{Model Interpretation with SHAP}
Interpretability is a crucial aspect of machine learning models, especially in domains where decision-making transparency is paramount, such as traffic accident injury fatality classification. To facilitate interpretability, the SHapley Additive exPlanations (SHAP) method was employed in this study. SHAP is a game-theoretic approach that assigns each feature an importance value indicating its contribution to the model's output \citep{lundberg_2017_a}. Unlike other interpretation techniques, SHAP provides both global and local explanations, allowing for a comprehensive understanding of the model's behavior. The basic methodology of SHAP involves computing Shapley values, which represent the average marginal contribution of a feature across all possible feature combinations. In the context of crash fatality classification, the calculation of SHAP values involves considering subsets of risk factors $S \subseteq F$, where $F$ denotes the set of all features influencing crash fatality. To quantify the impact of a specific factor $i$, two models are trained: one incorporating factor $i$, denoted as $f_{S\cup\{i\}}(x_{S\cup\{i\}})$, and another excluding factor $i$, represented by $f_{S}(x_{S})$, where $x_{S\cup\{i\}}$ and $x_{S}$ correspond to the input values of the risk factors. The difference in model outputs $f_{S\cup\{i\}}(x_{S\cup\{i\}}) - f_{S}(x_{S})$ is computed for every potential subset $S \subseteq F\backslash\{i\}$. The Shapley value of a risk factor $i$ is determined using Eq.~\ref{eq:shapley1}.

\begin{equation}
\label{eq:shapley1}
\phi_i = \sum_{S \subseteq F \setminus \{i\}} \frac{|S|!(|F| - |S| -1)!}{|F|!} \left [ f_{S \cup \{i\}}(x_{S \cup \{i\}}) - f_S(x_S) \right ].
\end{equation}

\section{Results and Discussion}

\subsection{Model Evaluation}
\label{evaluation}
\begin{figure}[!ht]
  \centering
  \includegraphics[width=0.46\textwidth]{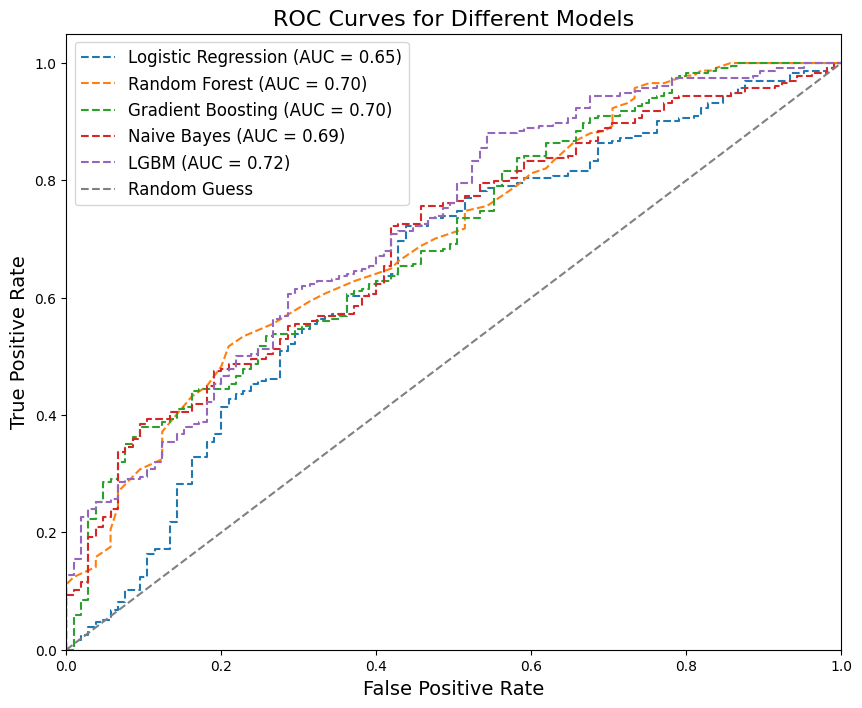}
  \caption{ROC-AUC curve of different ML models for traffic accident fatality prediction.}\label{fig:fig3}
\end{figure}

The performance of different machine learning classification algorithms for traffic accident fatality classification is assessed on the test set, which comprises 234 fatal and 114 non-fatal samples, as presented in Table~\ref{tab:model_evaluation}. The metrics assessed include precision, recall, mean cross-validation (CV) score, test accuracy, and F1 score. Additionally, the receiver operating characteristic (ROC) area under the curve (AUC) scores are provided in Figure~\ref{fig:fig3} for each model, offering insights into their discriminative capabilities.

Among the models evaluated, Light Gradient Boosting (LGBM) emerged as the optimal model, achieving the highest precision (0.72), recall (0.73), mean cross-validation score (0.76), test accuracy (0.74), F1-score (0.72), and ROC-AUC (0.72). This demonstrates LGBM's proficiency in accurately identifying cases with severe injuries while minimizing both false positives and false negatives. Random Forest Classifier and Gradient Boosting models also exhibit commendable performance, with comparable precision, recall, and test accuracy scores. However, they slightly underperform in terms of mean CV score compared to LGBM. While Logistic Regression and Naive Bayes algorithms demonstrate reasonable performance, they fall short compared to the ensemble-based models in terms of precision, recall, and F1 score. Although their test accuracy is competitive, their mean CV scores are relatively lower, indicating potential issues with generalization to unseen data. The superior results of LGBM likely stem from its tree-based structure, which effectively captures complex relationships in the data and incorporates inherent mechanisms to prevent overfitting. Employing a gradient boosting framework, LGBM prioritizes data partitioning based on the gradient of the loss function, utilizing a novel histogram-based approach for efficient and accurate feature discretization \citep{ke_2017_lightgbm}. Furthermore, LGBM implements a leaf-wise tree growth strategy, minimizing the loss function by greedily selecting the leaf with the maximum delta of the loss during each iteration, thereby enhancing predictive performance while reducing overfitting. However, the relatively modest AUC scores across all evaluated models signify the inherent complexity of this prediction task. This could warrant further exploration of more sophisticated feature engineering techniques, the collection of additional data, or the investigation of deep learning architectures specifically designed for complex pattern recognition.
\begin{table}[ht]
\caption{Performance evaluation of trained traffic accident fatality prediction models.}\label{tab:model_evaluation}
\begin{tabular*}{\hsize}{@{\extracolsep{\fill}}llllll@{}}
\toprule
\textbf{ML Classification Algorithms} & \textbf{Precision} & \textbf{Recall} & \textbf{Mean CV Score} & \textbf{Test Accuracy} & \textbf{F1 Score} \\
\midrule
Logistic Regression & 0.68 & 0.67 & 0.58 & 0.67 & 0.68 \\
Naive Bayes & 0.69 & 0.68 & 0.57 & 0.68 & 0.68 \\
Random Forest Classifier & 0.68 & 0.70 & \textbf{0.76} & 0.70 & 0.69 \\
Gradient Boosting & 0.68 & 0.70 & 0.73 & 0.69 & 0.69 \\
LightGBM & \textbf{0.72} & \textbf{0.73} & \textbf{0.76} & \textbf{0.74} & \textbf{0.72} \\
Multilayer Perception & 0.63 & 0.62 & 0.67 & 0.62 & 0.62 \\
\bottomrule
\end{tabular*}
\end{table}

\subsection{Global Model Explanation}
\begin{figure}[!ht]
    \centering
    \begin{minipage}[b]{0.43\textwidth}
        \centering
        \includegraphics[width=\textwidth]{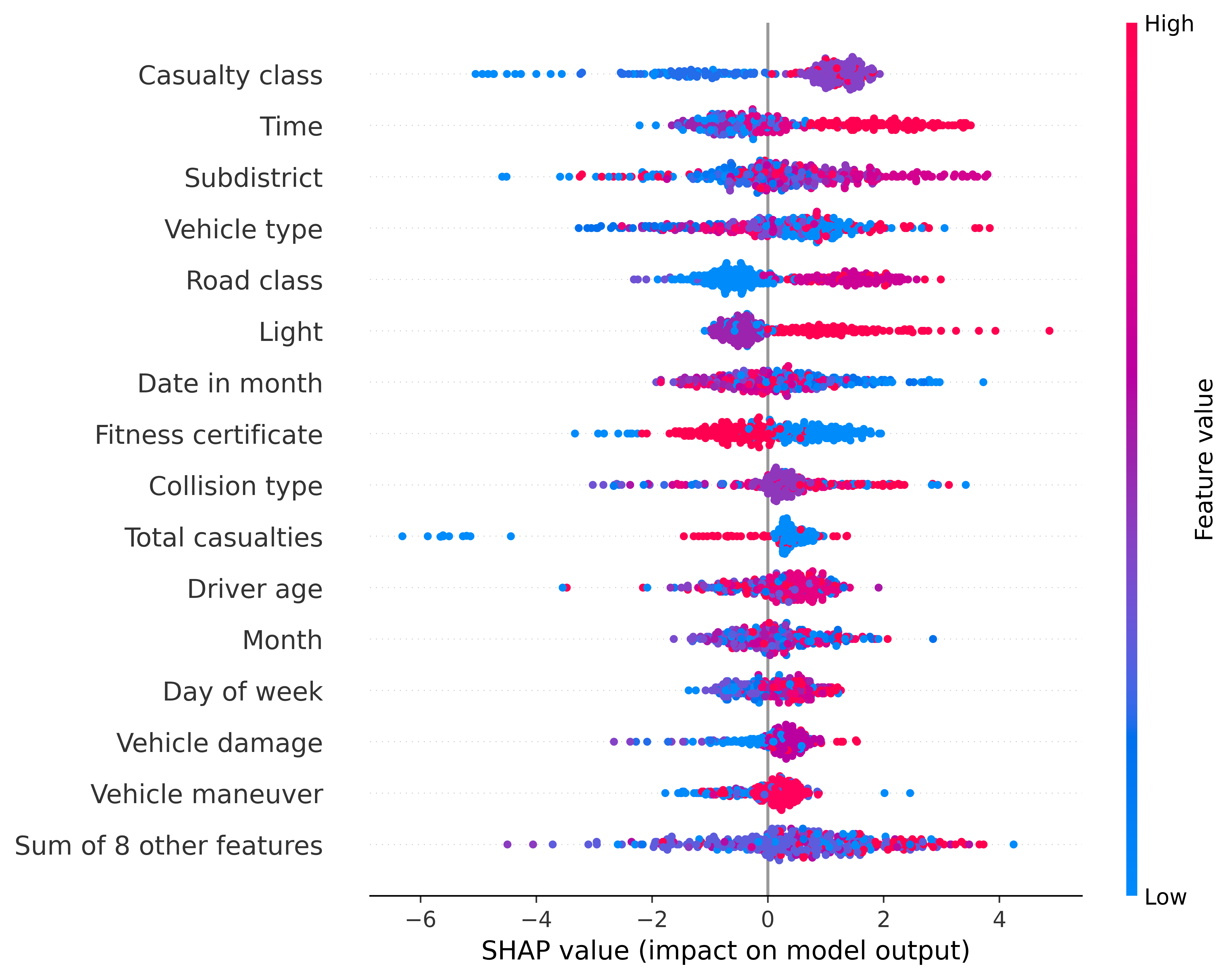}
    \end{minipage}
    \hfill
    \begin{minipage}[b]{0.56\textwidth}
        \centering
        \includegraphics[width=\textwidth]{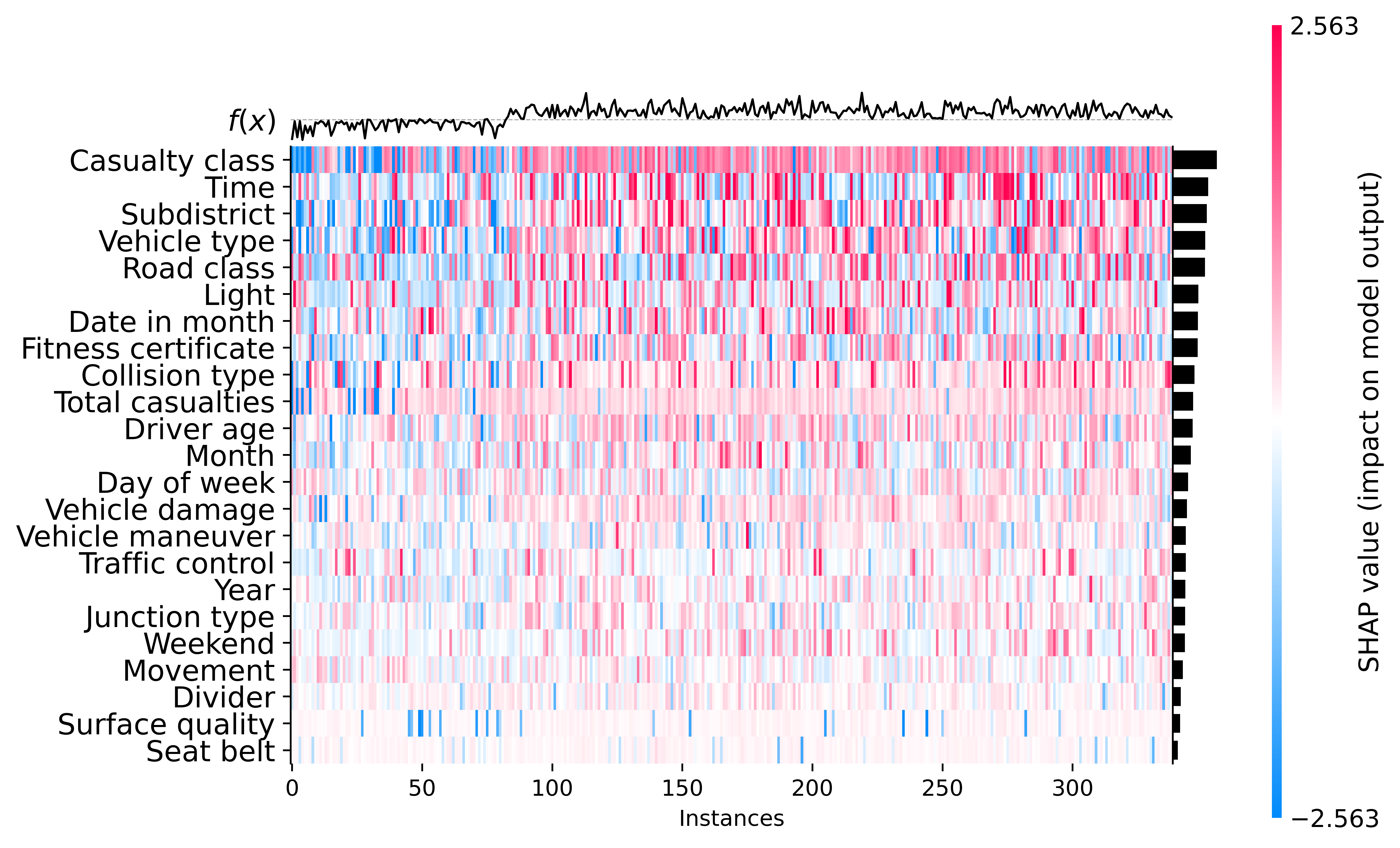}
    \end{minipage}
    \caption{(a) Beeswarm plot explaining LGBM on accident fatality; (b) Heatmap of SHAP values for all features across all samples in test set.}
    \label{fig:fig4}
\end{figure}

In Figure~\ref{fig:fig4}(a), The SHAP summary beeswarm plot illustrates the global significance of features in predicting traffic accident fatality. "Casualty Class", "Time", "Subdistrict", "Vehicle Type", and "Road Class" emerged as the top 5 influential factors. The features are listed on the y-axis by importance, with the most influential ones at the top, while the x-axis denotes SHAP values, indicating feature impact on log odds. Colors represent feature values, with red indicating high and blue indicating low values. Each point depicts an instance's contribution to predictions. For instance, "Road class" typically has a positive SHAP value when it is national or regional road, indicating national and regional highways propensity towards fatal injuries whereas city and feeder roads tend to be non-fatal for having negative SHAP values. Figure~\ref{fig:fig4}(b) depicts the SHAP heatmap, showcasing impact of each feature across all 375 test instances.  Individual instances are arranged on the x-axis, with features listed on the y-axis. The color of the line above each instance reflects its SHAP value for a particular feature, indicating whether that feature contributes positively towards fatal injury or negatively. The f(x) line illustrates model log odds output. The curve under the f(x) value demonstrates non-fatal cases and vice-versa. The bars on the right signify mean SHAP values, emphasizing overall feature importance.

\begin{figure}[!ht]
    \centering
    \begin{minipage}[b]{0.51\textwidth}
        \centering
        \includegraphics[width=\textwidth]{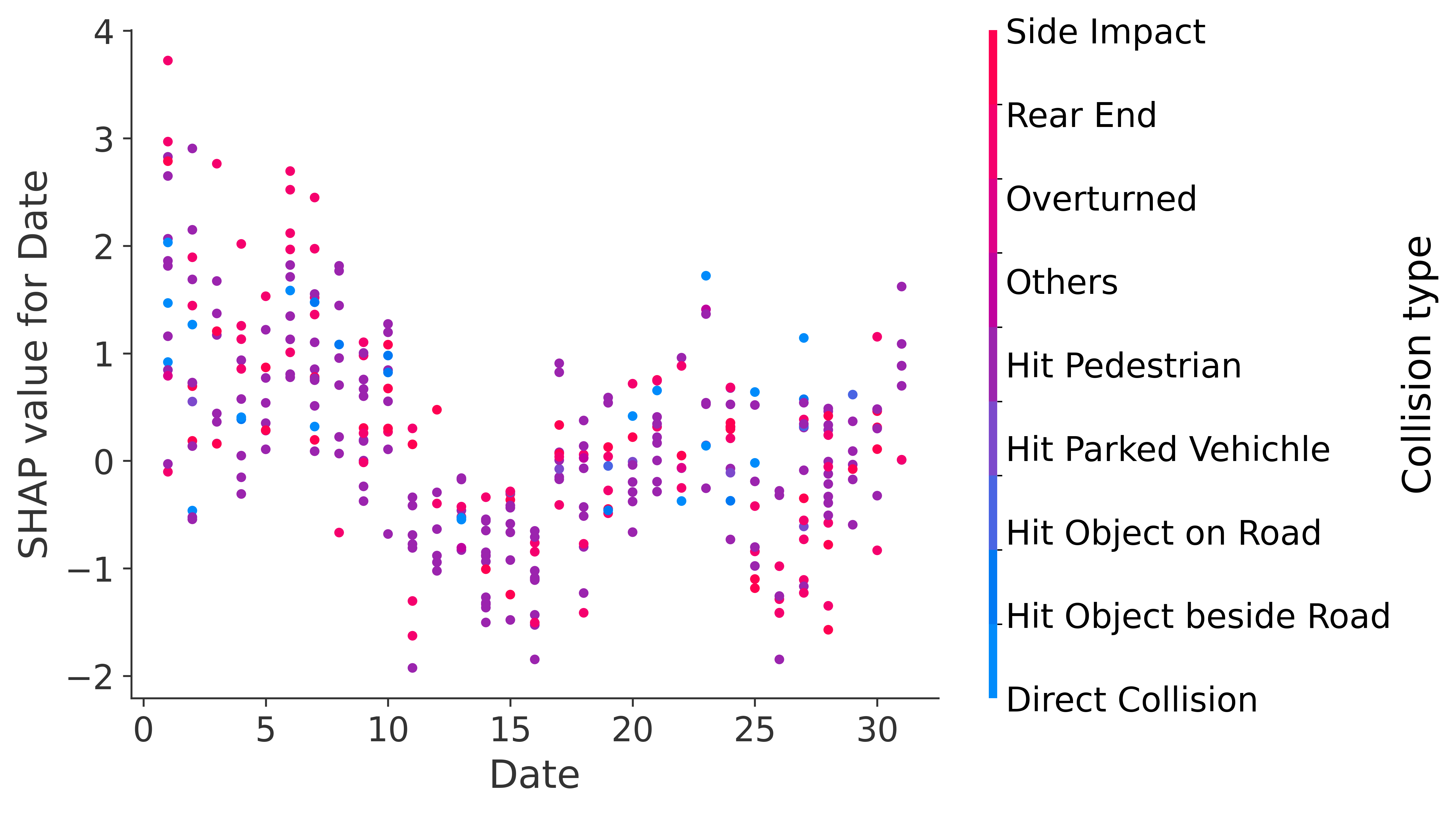}
    \end{minipage}
    \hfill
    \begin{minipage}[b]{0.48\textwidth}
        \centering
        \includegraphics[width=\textwidth]{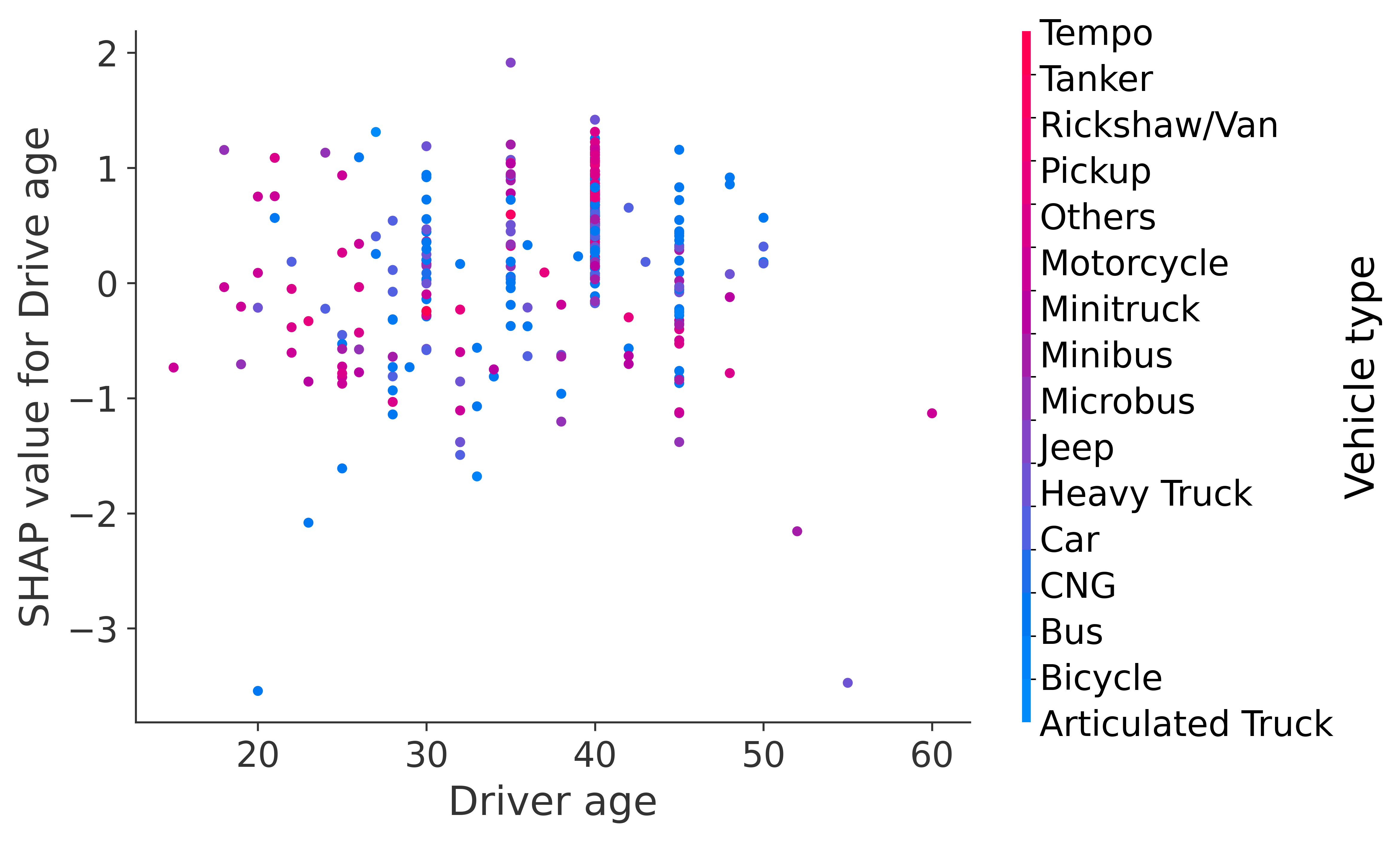}
    \end{minipage}
    \caption{Dependence plots (a) impact of Date and Collision type on model output; (b) impact of Driver age and Vehicle type on model output.}
    \label{fig:fig5}
\end{figure}

Although the beeswarm and heatmap plots offer a comprehensive view of SHAP values across multiple features simultaneously, a thorough understanding of the relationship between a feature's values and predicted accident fatality requires analysis of dependence plots. These plots reveal the marginal effect that features have on the target variable, accident fatality, and show the distribution of feature SHAP values. In a SHAP dependence scatter plot, the feature of interest appears on the horizontal axis, while its corresponding SHAP values are on the vertical axis. Each point on the scatter plot represents a data instance, displaying the feature's value and its associated SHAP value. The color of each point indicates a second feature that may interact with the feature of interest. By assigning the entire Explanation object to the color parameter, the scatter plot algorithm identifies the feature column with the strongest interaction. Figure~\ref{fig:fig5} In this analysis, we examine the influence of four key features: Date, Collision type, Driver age, and Vehicle type, as depicted in Figure~\ref{fig:fig5}. In Figure~\ref{fig:fig5}(a), we observe a non-linear relationship between date and corresponding SHAP values, along with their interaction with collision type. Notably, during the first week of each month, predominantly positive SHAP values are associated with rear-end and direct collisions, indicating a higher risk of fatal accidents. As the month progresses, these values decrease, suggesting a lower likelihood of fatal accidents. Among collision types, "Hit pedestrian," "rear end," "overturned," and "direct collision" emerge as the top four contributors to both fatal and non-fatal accidents. Figure~\ref{fig:fig5}(b) illustrates the interaction between driver age and vehicle type. It reveals that younger drivers under 30, particularly those riding motorcycles, are more prone to fatal accidents. In contrast, drivers aged between 30 to 50 tend to sustain injuries while driving buses, heavy trucks, and rickshaws. Additionally, older and more experienced drivers exhibit a lower probability of being involved in fatal accidents. However, a direct relationship between SHAP values and feature values is not evident from global interpretation due to differing feature relationships across predicted classes.
\subsection{Local Model Explanation}
\begin{figure}[!ht]
  \centering
  \includegraphics[width=0.8\textwidth]{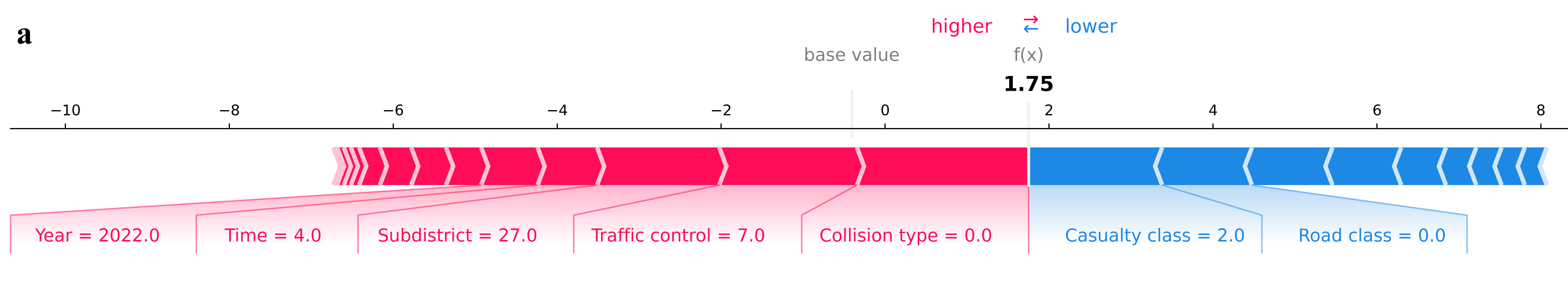}
  \includegraphics[width=0.8\textwidth]{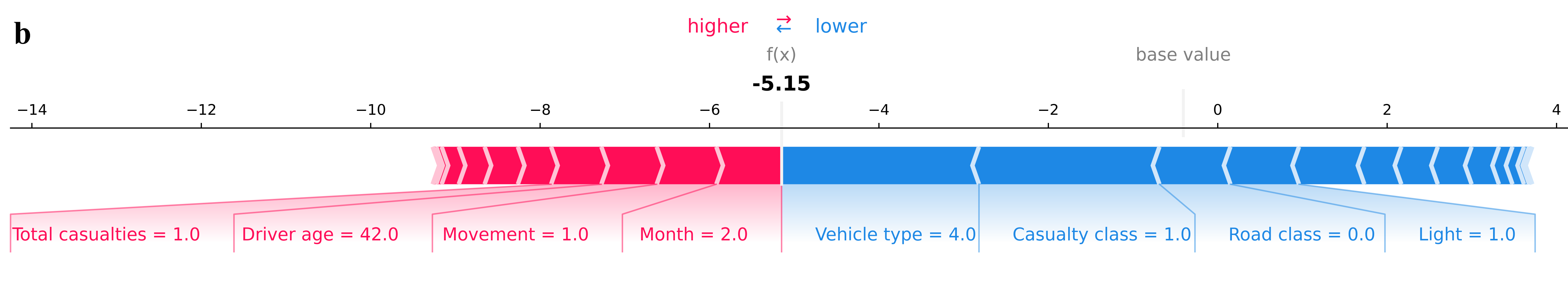}
  \caption{Force plots explaining individual sample of fatality prediction (a) fatal injury; (b) non-fatal injury.}\label{fig:fig6}
\end{figure}

To understand how each feature contributed to the fatality of individual cases, we have to look for local analysis. A SHAP force plot provides a detailed view of how individual features contribute to a model's prediction for a specific instance. Each feature's impact on the prediction is represented by a vertical bar, with the length indicating the magnitude and direction (positive or negative) of the effect. If we sum the SHAP values and the average prediction E[f(x)] we get the prediction for that instance f(x) in terms of log odds of a positive prediction as listed in Eq.~\ref{eq:shapley1}. In Figure~\ref{fig:fig6}(a), a test case is shown where an accident tends to be fatal. The predicted log odds of f(x)=1.75 exceeds the base value, E[f(x)] = -0.406. The top features affecting to be fatal in this scenario are direct collision, no traffic control, and late night driving. Figure~\ref{fig:fig6}(b) shows an observation where an accident tend to be non-fatal.
The model starts at the same base value of -0.406 and we can see how each feature contributed to the final prediction of -5.15. The major contributing factors towards non-fatal injury include the presence of light at night, city roads, and vehicles having fitness certificate.
\begin{equation}
\label{eq:shapley2}
f(x) = E[f(x)] + \mathrm{SHAP\ values}.
\end{equation}

\section{Conclusion and Future Work}
This study investigated the effectiveness of various machine learning algorithms for classifying traffic accident fatality with a focus on distinguishing between fatal and non-fatal outcomes. The major contributions include the development of predictive models using various machine learning algorithms and the application of the SHAP method for interpretability. Our meticulous data preprocessing, including SMOTE oversampling to address class imbalance, ensured a robust foundation for model development. The models achieved promising results, with the Light Gradient Boosting (LGBM) algorithm emerging as the optimal model, demonstrating high precision, recall, and overall predictive performance. It indicates suitability of tree-based methods for handling the complexities of accident data. Crucially, the integration of SHAP (SHapley Additive exPlanations) revealed significant insights into the influential factors affecting accident fatalities, such as casualty class, time, subdistrict, vehicle type, and road class. This transparency paves the way for evidence-based road safety interventions tailored to Bangladeshi conditions.

However, limitations include the inherent complexity of road accidents and the need for further exploration of sophisticated feature engineering techniques. Future work could involve incorporating more crash data and road safety attributes at the segment level to enhance model accuracy and capture additional insights into accident fatality. 

Overall, the insights gained from this research have practical implications for enhancing road safety and accident management strategies in developing countries. By effectively addressing the identified influential factors, authorities can deploy timely measures to mitigate the side effects of accidents and reduce casualties, ultimately leading to safer and more secure roadways.

\section{Acknowledgement}
This work was supported by Information and Communication Technology (ICT) Division under Ministry of Posts, Telecommunications and Information Technology, Government of the People's Republic of Bangladesh. Fellowship Grant No - 56.00.0000.052.33.001.23-09; Dated 04.02.2024.








\end{document}